\documentclass[10pt,twocolumn,letterpaper]{article}

\usepackage[pagenumbers]{cvpr} 

\usepackage{graphicx}
\usepackage{amsmath}
\usepackage{amssymb}
\usepackage{booktabs}

\usepackage[pagebackref,breaklinks,colorlinks]{hyperref}

\usepackage[capitalize]{cleveref}
\crefname{section}{Sec.}{Secs.}
\Crefname{section}{Section}{Sections}
\Crefname{table}{Table}{Tables}
\crefname{table}{Tab.}{Tabs.}

\def\cvprPaperID{*****} 
\def\confName{CVPR}
\def\confYear{2022}

\begin{document}

\title{Facial Action Unit Recognition Based on Transfer Learning}

\author{Shangfei Wang, Yanan Chang, Jiahe Wang\\
University of Science and Technology of China, Hefei, Anhui, China\\
{\tt\small sfwang@ustc.edu.cn, cyn123@mail.ustc.edu.cn, pia317@mail.ustc.edu.cn}
}
\maketitle

\begin{abstract}
    Facial action unit recognition is an important task for facial analysis. Owing to the complex collection environment, facial action unit recognition in the wild is still challenging. The 3rd competition on affective behavior analysis in-the-wild (ABAW) has provided large amount of facial images with facial action unit annotations. In this paper, we introduce a facial action unit recognition method based on transfer learning. We first use available facial images with expression labels to train the feature extraction network. Then we fine-tune the network for facial action unit recognition.
\end{abstract}

\section{Introduction}
\label{sec:intro}

Facial action units~(AU) are defined by facial action coding system~(FACS). We can describe facial expressions by the combination of AUs. Facial AU recognition has received a lot of attention due to its potential recently.

According to the collection environment, the databases with AU annotations can be divided into two categories, laboratory-controlled databases and in-the-wild databases. For example, BP4D~\cite{zhang2014bp4d} and DISFA~\cite{mavadati2013disfa} are two laboratory-controlled databases for AU recognition. Compared to laboratory-controlled databases, the collection environment of in-the-wild databases is more complex. The head pose, occlusion, and low resolution problems are common on the in-the-wild databases. Therefore, facial AU recognition on the in-the-wild condition is challenging. 

The 3rd competition on affective behavior analysis in-the-wild (ABAW)~\cite{kollias2022abaw, kollias2021analysing, kollias2020analysing, kollias2021distribution, kollias2021affect, kollias2019expression, kollias2019face, kollias2019deep, zafeiriou2017aff} has labeled lots of facial images with AU annotations. 547 videos of around 2.7M frames are included with 12 facial action units. According to previous works~\cite{jin2021multi}, the database also has data imbalance problem.

In order to improve facial AU recognition on the challenging database, we propose a method based on transfer learning. We first train the feature encoder on a balanced facial expression database. Then we fine-tune the model for facial AU recognition after adding AU classifiers. We also try to adjust the threshold of AUs to further enhance the performance.

\begin{figure*}[ht]
 
\centering
\includegraphics[scale=0.5]{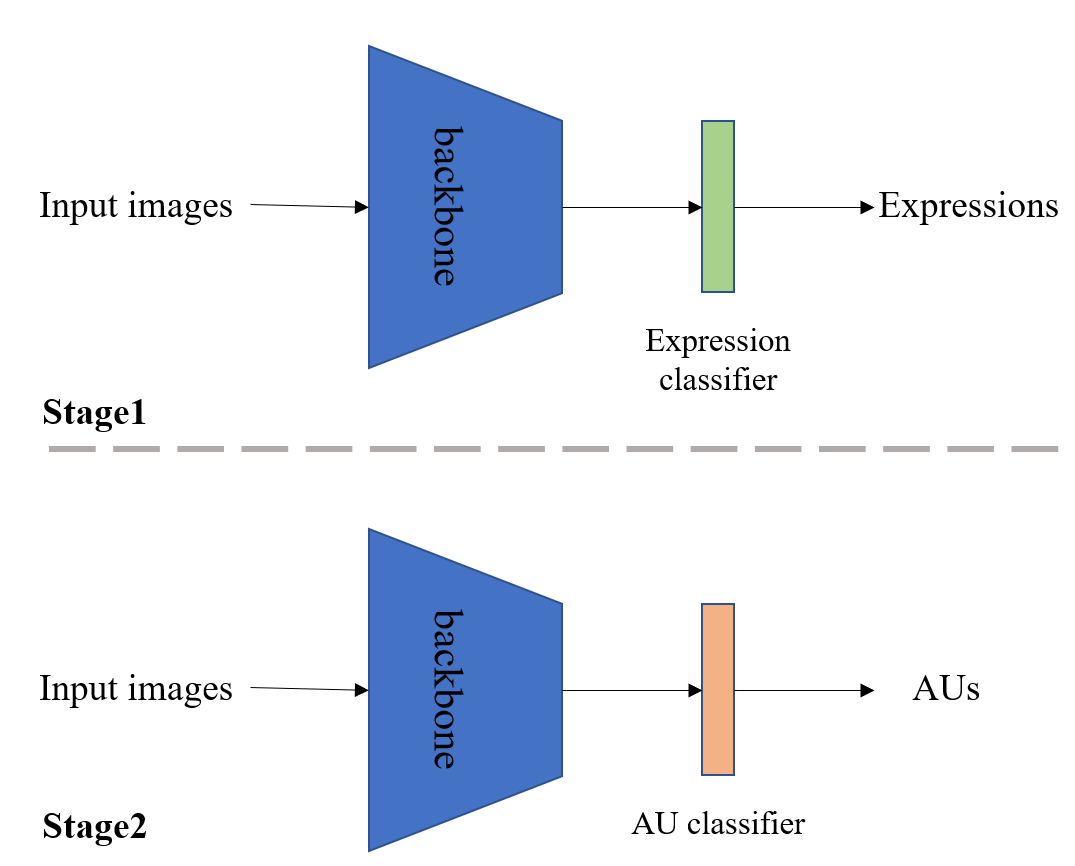}
\caption{The framework of the method. Two stages are involved. In stage one, the backbone network is trained by a facial expression recognition task. Based on the pre-trained feature extractor, AU classifier is trained for AU recognition in stage two.}
\label{framework}
 
\end{figure*}

\section{Method}
Figure~\ref{framework} shows the overall framework, which includes two stages. In the first stage, we train the feature extractor by a facial expression recognition task. In the second stage, we fine-tune the pre-trained feature extractor for AU recognition task. Through the two stage training, patterns learned from facial expression recognition can be transferred to facial AU recognition task. 

\subsection{Facial expression recognition task}
Facial expression recognition task tries to predict facial expressions. The overall network includes a backbone network to extract features from the input images and an expression classifier to predict expressions from the feature representations. The backbone is based on ResNet\_50 network~\cite{he2016deep}. The expression classifier is one linear layer. The training loss is cross-entropy loss. The loss is defined as follow:

\begin{equation}
  \mathcal{L}_{exp} = l_e(\hat{y}_e, y_e )
  \label{eq:1}
\end{equation}

where $l_e$ denotes cross-entropy loss, $\hat{y}_e$ and $y_e$ are the predicted and ground-truth facial expression labels, respectively.

\subsection{Facial AU recognition task}
After the training of stage one, we have a pre-trained backbone network. In order to transfer the learned patterns to facial AU recognition task, we train an AU classifier based on the pre-trained network. The AU classifier is a Multilayer Perceptron~(MLP), containing two hidden layers. The loss for AU recognition is defined as follow:

\begin{equation}
  \mathcal{L}_{au} = l_e(\hat{y}_a, y_a )
  \label{eq:2}
\end{equation}

where $l_e$ denotes cross-entropy loss, $\hat{y}_a$ and $y_a$ are the predicted and ground-truth AU labels, respectively. Through the two stage training, the patterns learned from facial expression task can be transferred to facial AU recognition. 

\section{Experiments}
\subsection{Databases}
Two databases are involved for the method, including Multi-PIE~\cite{gross2010multi} and Aff-Wild2~\cite{kollias2019expression}. The Multi-PIE database is used to train the facial expression recognition network. And the Aff-wild2 is used to train AU recognition network. For the training on stage one, we adopt five-fold subject-independent cross-validation. For stage two, we train the network on the training set. 

\subsection{Evaluation Metric}
For stage one, we adopt accuracy as the evaluation metric for facial expression recognition. For stage two, we evaluate the performance by F1 score for facial AU recognition. 

\subsection{Results}
For facial expression recognition in stage one, the average accuracy is 0.94. Multi-PIE database includes multi-view images. Through training the backbone network on the multi-view images, the backbone can learn pose-robust feature representations.

For facial AU recognition in stage two, the experimental result on the validation set is 0.460 F1 score. After choosing different thresholds for AU predicting, the best result for AU recognition is 0.47.

\section{Conclusion}

We propose a facial action unit recognition method based on transfer learning. The overall framework includes two stages. In the first stage, the backbone network is trained by a facial expression recognition task. In the second stage, we train AU classifier based on the pre-trained backbone network. The patterns learned from the well-trained backbone network on facial expression recognition task can benefit AU recognition.  

{\small
\bibliographystyle{ieee_fullname}
\bibliography{egbib}

\begin{thebibliography}{10}\itemsep=-1pt

\bibitem{gross2010multi}
Ralph Gross, Iain Matthews, Jeffrey Cohn, Takeo Kanade, and Simon Baker.
\newblock Multi-pie.
\newblock {\em Image and vision computing}, 28(5):807--813, 2010.

\bibitem{he2016deep}
Kaiming He, Xiangyu Zhang, Shaoqing Ren, and Jian Sun.
\newblock Deep residual learning for image recognition.
\newblock In {\em Proceedings of the IEEE conference on computer vision and
  pattern recognition}, pages 770--778, 2016.

\bibitem{jin2021multi}
Yue Jin, Tianqing Zheng, Chao Gao, and Guoqiang Xu.
\newblock A multi-modal and multi-task learning method for action unit and
  expression recognition.
\newblock {\em arXiv preprint arXiv:2107.04187}, 2021.

\bibitem{kollias2022abaw}
Dimitrios Kollias.
\newblock Abaw: Valence-arousal estimation, expression recognition, action unit
  detection \& multi-task learning challenges.
\newblock {\em arXiv preprint arXiv:2202.10659}, 2022.

\bibitem{kollias2020analysing}
D Kollias, A Schulc, E Hajiyev, and S Zafeiriou.
\newblock Analysing affective behavior in the first abaw 2020 competition.
\newblock In {\em 2020 15th IEEE International Conference on Automatic Face and
  Gesture Recognition (FG 2020)(FG)}, pages 794--800.

\bibitem{kollias2019face}
Dimitrios Kollias, Viktoriia Sharmanska, and Stefanos Zafeiriou.
\newblock Face behavior a la carte: Expressions, affect and action units in a
  single network.
\newblock {\em arXiv preprint arXiv:1910.11111}, 2019.

\bibitem{kollias2021distribution}
Dimitrios Kollias, Viktoriia Sharmanska, and Stefanos Zafeiriou.
\newblock Distribution matching for heterogeneous multi-task learning: a
  large-scale face study.
\newblock {\em arXiv preprint arXiv:2105.03790}, 2021.

\bibitem{kollias2019deep}
Dimitrios Kollias, Panagiotis Tzirakis, Mihalis~A Nicolaou, Athanasios
  Papaioannou, Guoying Zhao, Bj{\"o}rn Schuller, Irene Kotsia, and Stefanos
  Zafeiriou.
\newblock Deep affect prediction in-the-wild: Aff-wild database and challenge,
  deep architectures, and beyond.
\newblock {\em International Journal of Computer Vision}, pages 1--23, 2019.

\bibitem{kollias2019expression}
Dimitrios Kollias and Stefanos Zafeiriou.
\newblock Expression, affect, action unit recognition: Aff-wild2, multi-task
  learning and arcface.
\newblock {\em arXiv preprint arXiv:1910.04855}, 2019.

\bibitem{kollias2021affect}
Dimitrios Kollias and Stefanos Zafeiriou.
\newblock Affect analysis in-the-wild: Valence-arousal, expressions, action
  units and a unified framework.
\newblock {\em arXiv preprint arXiv:2103.15792}, 2021.

\bibitem{kollias2021analysing}
Dimitrios Kollias and Stefanos Zafeiriou.
\newblock Analysing affective behavior in the second abaw2 competition.
\newblock In {\em Proceedings of the IEEE/CVF International Conference on
  Computer Vision}, pages 3652--3660, 2021.

\bibitem{mavadati2013disfa}
S~Mohammad Mavadati, Mohammad~H Mahoor, Kevin Bartlett, Philip Trinh, and
  Jeffrey~F Cohn.
\newblock Disfa: A spontaneous facial action intensity database.
\newblock {\em IEEE Transactions on Affective Computing}, 4(2):151--160, 2013.

\bibitem{zafeiriou2017aff}
Stefanos Zafeiriou, Dimitrios Kollias, Mihalis~A Nicolaou, Athanasios
  Papaioannou, Guoying Zhao, and Irene Kotsia.
\newblock Aff-wild: Valence and arousal ‘in-the-wild’challenge.
\newblock In {\em Computer Vision and Pattern Recognition Workshops (CVPRW),
  2017 IEEE Conference on}, pages 1980--1987. IEEE, 2017.

\bibitem{zhang2014bp4d}
Xing Zhang, Lijun Yin, Jeffrey~F Cohn, Shaun Canavan, Michael Reale, Andy
  Horowitz, Peng Liu, and Jeffrey~M Girard.
\newblock Bp4d-spontaneous: a high-resolution spontaneous 3d dynamic facial
  expression database.
\newblock {\em Image and Vision Computing}, 32(10):692--706, 2014.

\end{thebibliography}
}

\end{document}